\documentclass[]{llncs}

\usepackage{wrapfig}
\usepackage{amsfonts}
\usepackage{mathtools}
\usepackage{amsmath}
\usepackage{amssymb}
\usepackage{cite}
\usepackage{array}
\usepackage{subfigure}
\usepackage{setspace}

\usepackage{graphicx}
\usepackage{booktabs}

\usepackage{multicol}
\usepackage{multirow}
\usepackage{float}
\usepackage[misc]{ifsym}
\newfloat{figtab}{htb}{fgtb}
\makeatletter
  \newcommand\figcaption{\def\@captype{figure}\caption}
  \newcommand\tabcaption{\def\@captype{table}\caption}
\makeatother

\newcommand{\eg}{\textit{e.g.}}
\newcommand{\ie}{\textit{i.e.}}
\newcommand{\etal}{\textit{et al.}}

\usepackage[pagebackref,breaklinks,colorlinks]{hyperref}

\usepackage{orcidlink}

\begin{document}

\title{Dealing with All-stage Missing Modality: Towards A Universal Model with Robust Reconstruction and Personalization}

\author{Yunpeng Zhao \inst{1}$^*$ \and
Cheng Chen \inst{2}$^*$ \and
Qing You Pang \inst{3} \and
Quanzheng Li \inst{2} \and \\
Carol Tang \inst{3,4} \and
Beng-Ti Ang \inst{3,4} \and
Yueming Jin \inst{1}\textsuperscript{\Letter}
}

\authorrunning{Y. Zhao et al.}

\institute{National University of Singapore \and
Harvard Medical School \& Massachusetts General Hospital \and
National Neuroscience Institute \and 
Duke-NUS Medical School\\
}

\maketitle

\def\thefootnote{*}\footnotetext{Equal Contribution. $\textsuperscript{\Letter}$Corresponding Author.}\def\thefootnote{\arabic{footnote}}

\begin{abstract}
  Addressing missing modalities presents a critical challenge in multimodal learning. Current approaches focus on developing models that can handle modality-incomplete inputs during inference, assuming that the full set of modalities are available for all the data during training. This reliance on full-modality data for training limits the use of abundant modality-incomplete samples that are often encountered in practical settings.
  In this paper, we propose a robust universal model with modality reconstruction and model personalization, which can effectively tackle the missing modality at both training and testing stages. 
  Our method leverages a multimodal masked autoencoder to reconstruct the missing modality and masked patches simultaneously, incorporating an innovative distribution approximation mechanism to fully utilize both modality-complete and modality-incomplete data. The reconstructed modalities then contributes to our designed data-model co-distillation scheme to guide the model learning in the presence of missing modalities. 
  Moreover, we propose a CLIP-driven hyper-network to personalize partial model parameters, enabling the model to adapt to each distinct missing modality scenario.
  Our method has been extensively validated on two brain tumor segmentation benchmarks. Experimental results demonstrate the promising performance of our method, which consistently exceeds previous state-of-the-art approaches under the all-stage missing modality settings with different missing ratios. Code will be available.
  
\end{abstract}

\section{Introduction}
\label{sec:intro}
Multimodal learning with multiple imaging modalities plays a crucial role in disease diagnosis and treatment planning.
Different MRI modalities, including T1-weighed (T1), post-contrast T1 (T1c), T2-weighted (T2), and Fluid Attenuated Inversion Recovery (FLAIR), are commonly-used to provide complementary information for accurate brain tumor segmentation \cite{ding2021rfnet, konwer2023enhancing}. However, in real-world clinical scenarios, the occurrence of missing one or several modalities is a prevalent situation due to artifacts, acquisition protocols, allergies to contrast agents, or economic considerations \cite{konwer2023enhancing, liu2023m3ae}. 
The missing modality problem can significantly hinder both the training and inference processes of conventional multimodal brain tumor segmentation methods that typically demand complete modality inputs. When encountering missing modality, these methods fail to make full use of modality-incomplete data during training and struggle to maintain performance during inference. Hence, addressing missing modalities effectively at both training and testing phases is crucial for robust multimodal learning.

\begin{figure}[t]
  \centering
  \includegraphics[width=0.9\linewidth]{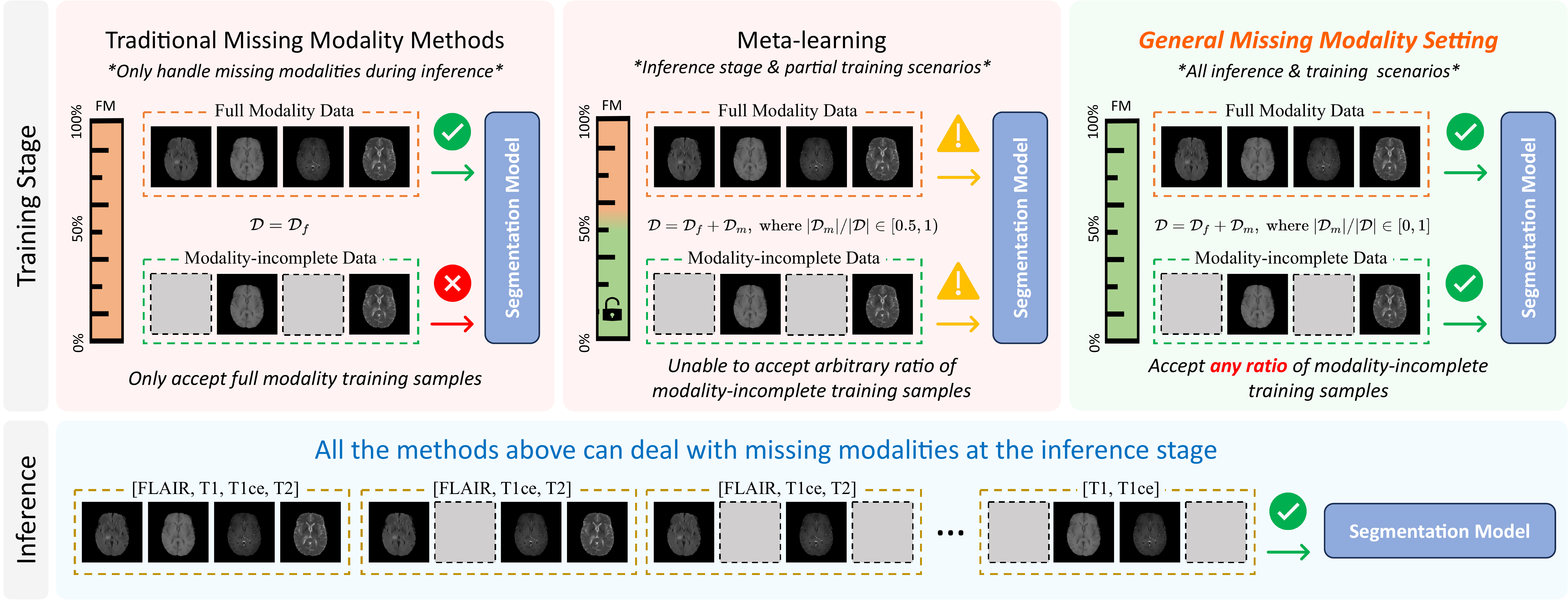}
  \caption{Our all-stage missing modality setting aims to address modality-incomplete input issues during both the training and inference phases simultaneously Moreover, we can also flexibly handle any ratio of full modality data within the train set. 
  }
  \label{fig:intro}
\end{figure}

Current approaches to addressing missing modality issues mainly focus on scenarios where modality-incomplete data is encountered during the inference stage, by leveraging strategies including shared latent spaces~\cite{ding2021rfnet, zhang2022mmformer}, knowledge distillation (KD)~\cite{hu2020knowledge, wang2021acn, azad2022smunet, liu2023m3ae}, and generative models~\cite{yang2023hypersyn, lee2020assessing}. However, these methods typically assume that training data include a complete set of modalities, which may be impractical since missing modalities can occur during both training and testing phases. As a result, such approaches are not optimized to fully leverage the valuable medical data with both modality-complete and modality-incomplete datasets for effective model training.
There are just very limited studies considering modality-incomplete data at both training and testing phases. SMIL~\cite{ma2021smil} adopted Bayesian meta-learning for missing modalities during training and testing on classification task. 
However, a limitation of this study is that it involves only two modalities, with missing data occurring in just one modality while the other remains consistently available. This constraint limits the flexibility compared to real-world scenarios, where both the number and type of missing modalities can be unpredictable.
Konwer~\etal~\cite{konwer2023enhancing} also developed a meta-learning framework for incomplete training and testing modalities, considering each missing modality combination as a distinct task for meta-training and meta-test. 
This method is constrained by the requirement that the proportion of modality-incomplete data during training cannot be too high or too low, to guarantee adequate full-modality data for meta-testing and sufficient missing-modality data for meta-training. This limitation restricts its adaptability in scenarios with varying ratios of missing modality data.

In this paper, we aim to tackle \textbf{all-stage missing modality} by developing a universal model capable of utilizing varying ratios of modality-complete and modality-incomplete data during training, and producing robust predictions for various missing modality scenarios during testing. To achieve this, two key challenges must be overcome. \textbf{First}, the loss of critical information due to missing modalities is an obvious issue, one that has received considerable attention in prior research. However, our challenge extends beyond just the testing phase; we also face information loss during training. This dual-phase information loss imposes more stringent demands and challenges on our method's design.
\textbf{Second}, the varying missing modality combinations during training and testing represents a significant challenge due to the highly heterogeneous data distributions, a factor often overlooked in previous works on missing modality. Taking brain tumor segmentation as an example, four modalities can result in fifteen different missing modality combinations, corresponding to fifteen distinct data distributions. Forcing a model to process these diverse distributions with shared parameters has been shown to be suboptimal in prior multimodal learning studies~\cite{dou2020unpaired}.   

To address these two critical challenges in all-stage missing modality, we propose a universal model with robust modality reconstruction and model personalization. 
Regarding the issue of information loss, one intuitive approach is to leverage the available modalities to reconstruct the missing ones. In this process, the model learns to exploit the inherent inter-modal correlations to infer the missing information. Concurrently, the reconstructed modalities can be integrated with the available ones to form a complete set of modalities. Drawing inspiration from the multimodal masked autoencoder \cite{liu2023m3ae}, we pre-train our model to simultaneously reconstruct the missing modality and masked image patches, but we further innovate a distribution approximation mechanism that allows us to fully utilize a variety of modality-complete and modality-incomplete data for modality reconstruction. Once the modality is reconstructed, we propose a data-model co-distillation scheme to use the reconstructed full modality information to guide the learning of the model with missing modalities. To tackle the issue of data distribution heterogeneity caused by different missing modalities, we consider a possible solution could be to enable the model personalize a small portion of parameters to accommodate each unique missing modality scenario. We propose a CLIP-driven hyper-network to personalize partial model parameters, which combines textual modality prompts with visual embeddings to serve as informative indicators. Our contribution can be summarized as follows:

\begin{itemize}
    \item We propose a robust universal model to effectively deal with missing modality during both training and testing phases. To the best of our knowledge, this is the first work that tackles the challenging all-stage missing modality scenario with the flexibility of accommodating every possible type and proportion of absent modalities.
    \item We innovate a distribution approximation mechanism for modality reconstruction with a multimodal masked autoencoder, which can learn the inter-modal correlations with modality-incomplete training data.
    \item We devise a data-model co-distillation scheme to effectively utilize the reconstructed modality information for guiding the model training with incomplete training modalities.
    \item We design a CLIP-driven hyper-network to provide personalized prediction-level parameters according to the textual prompt indicating the input modalities accommodate each unique missing modality scenario.
    \item Extensive experiments are conducted on two brain tumor segmentation benchmarks, proving the superiority of our method under the all-stage missing modality settings. With only 1\% full modality in training data, our model can achieve superior results than the others with 100\% ratio.
\end{itemize}

\section{Related Work}
\label{sec:related_work}

\textbf{Brain tumor segmentation with missing modalities.} 
Multimodal learning for brain tumor segmentation often faces challenges with modality-incomplete data, which will lead to notable declines in performance. Currently, numerous of efforts \cite{ding2021rfnet, liu2023m3ae, zhang2022mmformer, wang2021acn, azad2022smunet, wang2023shaspec, dorent2019hved, havaei2016hemis}, namely missing modality methods, have been designed to learn a model that is robust to partial modality inputs. Some approaches focused on knowledge distillation \cite{hu2020knowledge, wei2023mmanet, wang2021acn, azad2022smunet}, aiming to facilitate the transfer of knowledge from a teacher network, trained with full modality data, to a student model that lacks one or more modalities. 
Another popular solution is shared latent space models \cite{ding2021rfnet, zhang2022mmformer}, which attempt to encode different modalities into a common latent embedding subspace.
Additionally, several works also counted on generative methods \cite{yang2023hypersyn, lee2020assessing} to synthesize missing modalities using the available modalities, and employ the completed full-modality data for segmentation. However, all these methods require the modality completeness of every training sample, which means that they fail to deal with missing modality issue during training. This could be unrealistic because in real clinical studies, full modality data is usually extremely limited and valuable due to various factors such as motion artifacts, acquisition settings, and even budget constraints. In contrast, modality-incomplete data is often sufficient and can be more accessible. Hence, we focus on a general missing modality setting that aim to deal with modality-incomplete data during both training and inference. 

\noindent\textbf{Missing modalities at training stage.} 
Although Konwer \etal \cite{konwer2023enhancing} took a step to utilize both full modality data and modality-incomplete data during training, their meta-learning framework lacks flexibility and does not permit varying proportions of full modality data. It will become ineffective with high full modality ratio in the train set. Furthermore, they failed to recover additional missing modality information in the training set, hindering their efficient utilization of inter-modal relationships present in the existing data. Particularly, when the missing modality problem during the training phase becomes severe (\ie, $|\mathcal{D}_f| \ll |\mathcal{D}_m|$), their performance also declines significantly.

\section{Methodology}
\label{sec:method}

\begin{figure}[t]
  \centering
  \includegraphics[width=1.0\linewidth]{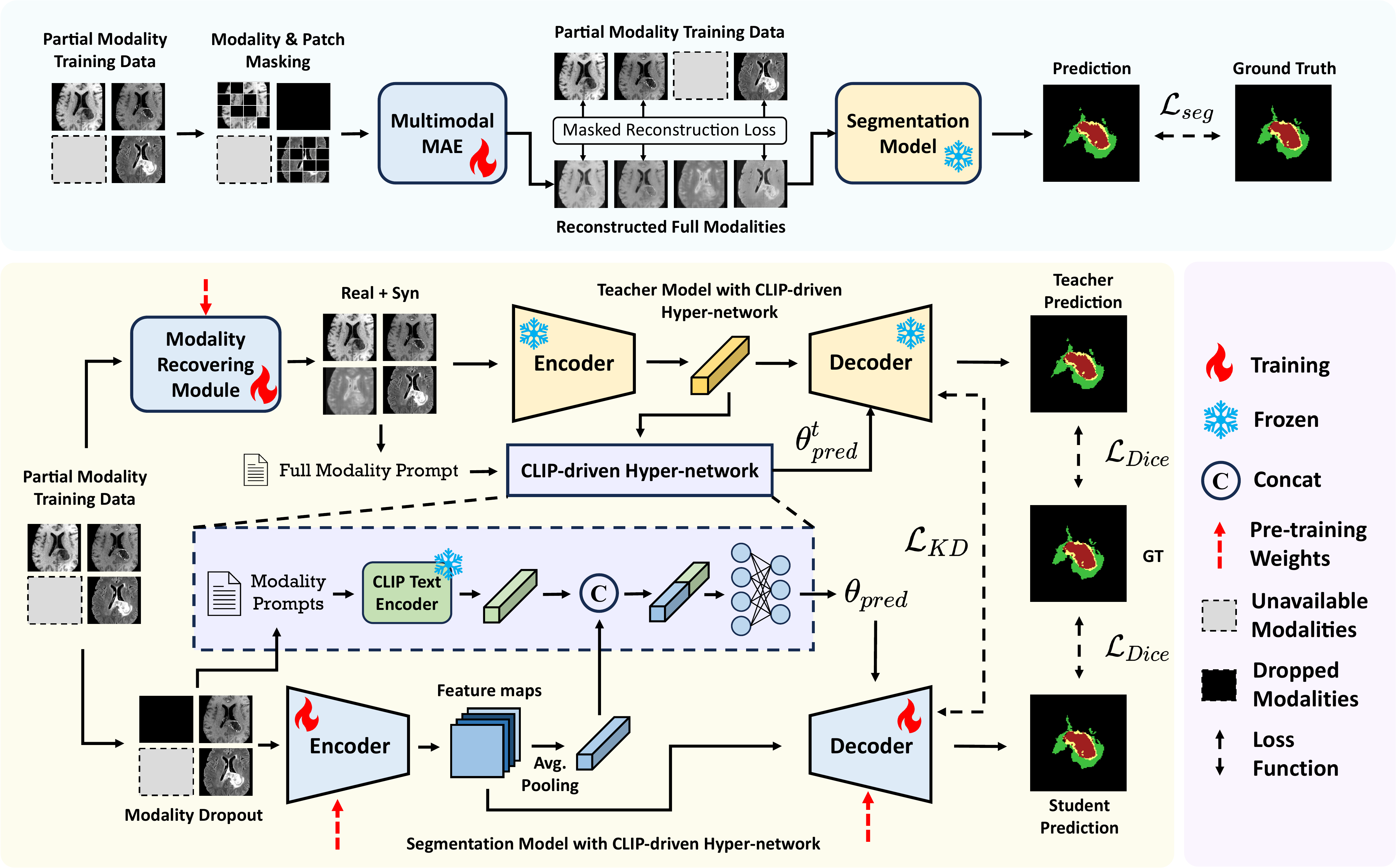}
  \caption{Overview of our proposed universal model for dealing with all-stage missing modality. We first propose a distribution approximation mechanism for a robust modality reconstruction (blue). The data-model co-distillation scheme is then designed to use reconstructed full modality to guide the model learning, in which CLIP-driven hyper-network is proposed to tackle distribution heterogeneity (Green).
  }
  \label{fig:framework}
\end{figure}

In this section, we introduce our universal multimodal brain tumor segmentation method for handling all-stage missing modalities. 
The whole framework is shown in Fig.~\ref{fig:framework}. We first proposed a novel partial-modality pre-training strategy via distribution approximation to train a multimodal MAE, in order to learn robust multimodal brain tumor representations with incomplete training modalities. 
Afterwards, the learned parameters are loaded onto our all-stage missing modality segmentation model as initial weights, as they can be a promising starting point for optimization.
During the training process of the segmentation model, we develop a data-model co-distillation scheme. By jointly refining the modality completion process and performing knowledge distillation, we can employ the synthesized full modality representations within the teacher model to assist model learning. 
Moreover, our CLIP-driven hyper-network is integrated into the teacher model and the segmentation model. It enables the model to flexibly adapt to varying missing modality scenarios, to deal with the heterogeneous data distribution during training and testing.
At the inference phase, we only use the segmentation model equipped with CLIP-driven hyper-network, and modality-incomplete data can be fed into our model to generate robust segmentation results.

\subsection{Modality Reconstruction via Distribution Approximation}
\label{ssec:pre}

In addressing the challenge of incomplete training modalities, a crucial aspect is learning inter-modal correlations using existing data. We have identified self-supervised learning as an exceptionally suitable solution, where, through learning by reconstructing, the model learns how to recover the information of missing modalities from the available ones.
To this end, we build a multimodal MAE that shares the same network architecture as the segmentation model we aim to train, with the exception that the final layer is replaced by an $M$-channel $1 \times 1$ convolutional layer. $M$ is the total number of modalities. Modality dropout and patch-wise masking strategy are performed on training samples to construct the incomplete inputs. Then the multimodal MAE will be trained to reconstruct the corresponding full modalities from the partial observation. It is obvious that the training process require complete modalities as ground truth for self-supervision. However, in our all-stage missing modality setting, the train set $\mathcal{D} = \{ \mathcal{D}_f, \mathcal{D}_m \}$, where $\mathcal{D}_f$ and $\mathcal{D}_m$ denotes the subset of full modality data and modality-incomplete data respectively.
Therefore, for a training sample $\boldsymbol{x} \in \mathcal{D}_m$, where $\boldsymbol{x} = \{x_i\}_{i=1}^M$ and $x_i = \boldsymbol{0}$ if the $i$-th modality is missing, only those available modalities can be used to calculate the reconstruction loss:
\begin{equation}
    \mathcal{L}_{\mathit{rec}}(\hat{\boldsymbol{x}}, \boldsymbol{x}) = \frac{1}{|\{x_i| x_i \neq \mathbf{0}\}|} \sum_{i \in \{i| x_i \neq \mathbf{0}\}} \| {\hat x}_i - x_i \|^2,
    \label{eq:rec_loss}
\end{equation}
where $\hat{\boldsymbol{x}}$ is the reconstructed full modalities. 
The incomplete self-supervision will lead to suboptimal pre-training weights as the reconstruction process of the absent modalities is uncontrollable. To provide auxiliary supervision for modalities without ground truth, we propose a partial modality pre-training strategy via distribution approximation.

Assuming the presence of a segmentation model trained on $\mathcal{D}$ with favorable generalization capabilities, it should exhibit superior performance (\ie,~the lowest segmentation loss) when provided with full modality inputs compared to any other modality combination. Empirically, by fixing the parameters of the segmentation model and minimizing the segmentation loss to optimize the inputs, we can maximize the probability that the input modalities conform to the full modality distribution. The idea is that a lower loss usually indicates a higher likelihood that the given segmentation model perceives the input data as being a full modality sample. This process can be formulated as:
\begin{equation}
    \hat{\boldsymbol{x}}^* = \mathop{\arg\min}\limits_{\hat{\boldsymbol{x}}} \mathcal{L}_{\mathit{seg}}(f(\hat{\boldsymbol{x}}), \boldsymbol{y}),
    \label{eq:distribution_approximation}
\end{equation}
where $\boldsymbol{y}$ is the segmentation label, $f(\cdot)$ denotes a segmentation model, and $\mathcal{L}_{\mathit{seg}}$ is the segmentation loss, \eg,~cross-entropy. In practice, we can use pre-trained parameters from Eq.~(\ref{eq:rec_loss}) to initialize a segmentation model, and train it on $\mathcal{D}$ to obtain $f(\cdot)$.
It can be seen from Eq.~(\ref{eq:distribution_approximation}) that by approximating $\hat{\boldsymbol{x}}$ towards full modality distribution, we provide extra supervision for all the reconstructed modalities, including those with absent self-supervision. Noting that $\hat{\boldsymbol{x}}$ is determined by the parameters of the MAE, our partial-modality pre-training process can be represented as:
\begin{equation}
    \min_\theta \mathcal{L}_{\mathit{rec}}(\phi_\theta(\mathtt{MASK}(\boldsymbol{x})), \boldsymbol{x}) + \lambda \mathcal{L}_{\mathit{seg}}(f(\phi_\theta(\mathtt{MASK}(\boldsymbol{x}))), \boldsymbol{y}),
    \label{eq:pre}
\end{equation}
where $\phi_\theta(\cdot)$ denotes MAE parameterized by $\theta$, $\mathtt{MASK}(\cdot)$ is our masking strategy, and $\lambda$ is a factor set to 0.1 that controls the supervision for absent modalities. 
Essentially, our proposed distribution approximation can not only deal with modality-incomplete training data, but also generally improves the quality of pre-training weights by implicitly guiding the model to learn tumor-aware multimodal representations. When minimizing the segmentation loss in Eq.~(\ref{eq:distribution_approximation}), the pre-training process is forced to learn tumor-centric information contributing to better segmentation accuracy, instead of over-reconstructing those irrelevant voxels, especially the black areas in the background (see results in Fig.~\ref{fig:vis_inv}).

\subsection{Data-model Co-distillation with Reconstructed Modality}
\label{ssec:data_model}

After obtaining the pre-trained MAE with inter-modal knowledge, we can leverage it to synthesize the missing training modalities, thereby recovering incompleteness and easing the challenge of information loss.
However, we observe a discrepancy between synthesized modalities and their real counterparts (see Fig.~\ref{fig:vis_inv}). Directly using low-quality synthetic data to train models may lead to noise interference problems. 
To address this issue, we propose a data-model co-distillation scheme that effectively utilizes the information from modalities synthesized by MAE.
The loss of information due to incomplete training modalities therefore can be propagated and compensated by the designed distillation scheme.

As shown in Fig.~\ref{fig:framework}, our data-model co-distillation scheme consists of a learnable modality recovering module, a frozen teacher model and to-be-learned segmentation model as student. 
We initialize teacher model, similar to the segmentation model in Sec.~\ref{ssec:pre}.
We utilize the pre-trained MAE as the initialization of the modality recovering module, to synthesize the missing modality information during training.
For $\boldsymbol{x} \in \mathcal{D}_m$, we generate synthetic full modalities $\hat{\boldsymbol{x}}$. 
For those absent modalities in $\boldsymbol{x}$, we fill in the blanks using the corresponding reconstructed images from $\hat{\boldsymbol{x}}$ to obtain the recovered `full modality' data $\hat{\boldsymbol{x}}_{\mathit{f}}$, which are then fed into the teacher model.
Meanwhile, the input to the segmentation student model is $\boldsymbol{x}$ after modality dropout.
To avoid suboptimal performance due to domain gap, our model distillation implicitly propagates the information of synthesized modalities from the teacher model to the student segmentation model. The general knowledge distillation is used and its object as follows:
\begin{equation}
    \mathcal{L}_{\mathit{KD}} = \sum_{l=1}^{L} \mathcal{L}_{\mathit{MAD}} (\boldsymbol{f}_l^{\phi_t}, \boldsymbol{f}_l^{\phi_s}),
    \label{eq:kd}
\end{equation}
where $\boldsymbol{f}_l^{\phi_t}$ and $\boldsymbol{f}_l^{\phi_s}$ are the feature maps at the $l$-th layer of teacher model $\phi_t$ and our segmentation model ${\phi_s}$, $L$ is the total number of layers for KD, and $\mathcal{L}_{\mathit{MAD}}$ denotes the margin-aware distillation loss for segmentation \cite{wei2023mmanet}. 
Through performing KD, we enhance partial modality representation $\boldsymbol{f}_l^{\phi_s}$ to a more informative representation, benefiting from the knowledge of synthetic modalities.

Note that the modality recovering module is learnable during training along with the model distillation, it means that we also refine the generation process of synthesized modalities through distribution approximation:
\begin{equation}
    \mathcal{L}_{\mathit{data}} = \mathcal{L}_{\mathit{seg}}(\phi_t(\hat{\boldsymbol{x}}_f), \boldsymbol{y}),
    \label{eq:datadist}
\end{equation}
where $\mathcal{L}_{\mathit{seg}}$ is Dice loss, and $\boldsymbol{y}$ is  segmentation label. Since the parameters of $\phi_t$ are frozen, $\mathcal{L}_{\mathit{data}}$ is only for refining the modality recovering process.
The generation of missing modalities in training data can be benefitted in turn (data distillation).
With our data-model co-distillation, we can effectively leverage the knowledge from recovered modality to benefit the segmentation model training.

\subsection{CLIP-driven Hyper-network for Model Personalization}
\label{ssec:clip_hyper}

Beyond information loss, another challenge posed by missing modalities is the significant heterogeneity in modality distributions. We propose a possible solution that involves using a hyper-network to adaptively adjust a subset of model parameters for heterogeneous data. 
Hyper-network architecture \cite{ha2017hypernetworks} involves using a simple neural network, to generate parameters for another primary network, which allows the primary network to adapt its weights on the fly to different tasks or data points. Consequently, by providing appropriate modality indicators to the hyper-network, we can generate personalized parameters at the prediction-level for better adaption to heterogeneous missing modality scenarios during both training and inference, thereby benefiting our all-stage setting. 

The inputs of a segmentation model equipped with our CLIP-driven hyper-network will be images $\boldsymbol{x} = \{x_i\}_{i=1}^M$, along with a binary modality code $\boldsymbol{c} = \{c_i | c_i = 1\ \text{if}\ x_i \neq \boldsymbol{0}\ \text{else}\ 0\}_{i=1}^{M}$ indicating the availability of modalities. Then we can formulate a textual prompt according to $\boldsymbol{c}$, which will be fed into the pre-trained large-scale CLIP model to generate textual embedding $\boldsymbol{Z}_{m}$. The template of textual prompt is ``\textit{The input MRI modalities are \{available modalities\}}". Upon obtaining $\boldsymbol{Z}_{m}$, we apply global average pooling to the feature maps outputted by the encoder to derive a feature-level representation vector $\boldsymbol{Z}_{\mathit{e}}$. Subsequently, the text prompt embedding is prepended to the visual embedding to form the hyper-network indicator. Hence, the customized prediction-level parameters are generated by 
\begin{equation}
    \{ \boldsymbol{w}_i, \boldsymbol{b}_i \}_{i=1}^{L_{\mathit{hyp}}} = \mathtt{MLP}(\boldsymbol{Z}_{m} \oplus \boldsymbol{Z}_{e}),    
\end{equation}
where $\boldsymbol{w}_i$ and $\boldsymbol{b}_i$ denotes weights and bias of the $i$-th layer, and $L_{\mathit{hyp}}$ is the total number of layers affected by our CLIP-driven hyper-network.
Through concurrently considering both textual modality descriptions and visual features, our hyper-network modifies certain model parameters based on the modality and content of different input data, thereby offering a wealth of customized representations within a unified framework.
Additionally, the reason why we use $\boldsymbol{Z}_{m}$ rather than $\boldsymbol{c}$ to form the indicator lies in the assumption of modality orthogonality implicit in simple binary modality codes, whereas the CLIP text encoder is capable of modeling the semantic correlations between different MRI modalities. More informative indicators can benefit the quality of parameter generation by the hyper-network. 

An interesting finding is that the hyper-teacher, \ie, the teacher model in Sec.~\ref{ssec:data_model} but also equipped with a hyper-network, is a magic key to the effectiveness of data-model co-distillation. 
By setting the input modality code to all-ones, the parameters of the hyper-teacher are locked in a full modality mode. Therefore, this can ensure that we are leveraging the full modality knowledge within the hyper-teacher, rather than other similar modality combinations, to benefit both the synthetic modalities refining and the knowledge transfer process.

\subsection{Overall Objective for Fine-tuning}

For our segmentation network, the Dice loss is used to minimize the difference between the predicted segmentation mask and the ground truth, which will be $\mathcal{L}_{\mathit{task}} = \mathcal{L}_{\mathit{dice}}(\phi_s(\mathtt{MDrop}(\boldsymbol{x}), \boldsymbol{c}), \boldsymbol{y})$, where $\mathtt{MDrop}(\cdot)$ denotes the random modality drop operator and $\boldsymbol{c}$ is the modality code describing available modalities for CLIP-driven hyper-network. We jointly utilize segmentation loss, data refining loss in Eq.~(\ref{eq:datadist}), and KD loss in Eq.~(\ref{eq:kd}) to optimize our framework.
\begin{equation}
    \min_{\theta_{\mathit{seg}}, \theta_g} \mathcal{L}_{\mathit{task}} + \alpha \mathcal{L}_{\mathit{data}} + \beta \mathcal{L}_{\mathit{KD}},
\end{equation}
where $\theta_{\mathit{seg}}$ represents the parameters of our segmentation model, $\theta_g$ is the parameters of the learnable modality recovering module, $\alpha$ and $\beta$ are used to control the data refining process and knowledge distillation respectively. Empirically, $\alpha$ and $\beta$ are set to 1.0 and 0.1.

\section{Experiment}
\label{sec:exps}

\subsection{Datasets and Metrics}
\label{ssec:dataset}

\textbf{Datasets.} We evaluate our method on two widely-used brain tumor segmentation benchmarks from BraTS2018 and BraTS2020 challenges~\cite{menze2014brats}. They comprise 285 and 369 preprocessed training cases in size of $155 \times 240 \times 240$ with ground truth, respectively, and each subject has complete four MR sequences. The segmentation labels include three classes, which are whole tumor (WT), tumor core (TC), and enhancing tumor (ET). We follow the data split in previous works~\cite{liu2023m3ae, ding2021rfnet} for training, validation and testing, which will be 199:29:57 for BraTS2018 and 219:50:100 for BraTS2020. For the preprocessing, all of the MRI images in these datasets have been skull-stripped, co-registered to the same template, and re-sampled to 1 mm$^3$ resolution by the challenge organizer.

\noindent\textbf{Evaluation Metrics.} Following the common practice for BraTS datasets, we adopt the Dice similarity coefficient (DSC $\uparrow$) and the 95th percentile Hausdorff distance (HD95 $\downarrow$) to quantitatively evaluate the segmentation performance.

\begin{table}[ht]
\centering
\caption{Comparison with state-of-the-art on segmentation of tumor regions (ET, TC, WT) under different FM ratios (1\%, 10\%, and 50\%) on BraTS2018. DSC across 15 input modality scenarios are reported.}
\resizebox{\textwidth}{!}{
\begin{tabular}{c|c|l|c|c|c|c|c|c|c|c|c|c|c|c|c|c|c|c}
\hline\hline
\multicolumn{1}{c|}{\multirow{4}{*}{Ratio}} & \multicolumn{1}{c|}{\multirow{4}{*}{Area}} & FLAIR & $\circ$ & $\circ$ & $\circ$ & $\bullet$ & $\circ$  & $\circ$  & $\bullet$ & $\circ$ & $\bullet$ & $\bullet$ &$\bullet$  & $\bullet$ & $\bullet$ & $\circ$ & $\bullet$ & \multirow{4}{*}{\begin{tabular}[c]{@{}l@{}} Mean \end{tabular}} \\ 
\multicolumn{1}{c|}{} & \multicolumn{1}{c|}{}                  & T1    &  $\circ$ & $\circ$  &$\bullet$  & $\circ$ & $\circ$ & $\bullet$ & $\bullet$ & $\bullet$ & $\circ$ & $\circ$ & $\bullet$ & $\bullet$ & $\circ$  & $\bullet$ & $\bullet$ &                           \\ 
\multicolumn{1}{c|}{} & \multicolumn{1}{c|}{}                   & T1c  & $\circ$ &$\bullet$ & $\circ$ & $\circ$ &$\bullet$  &$\bullet$  & $\circ$ & $\circ$ & $\circ$ & $\bullet$ & $\bullet$ & $\circ$ & $\bullet$ & $\bullet$ & $\bullet$ &                             \\
\multicolumn{1}{c|}{} & \multicolumn{1}{c|}{}                   & T2    & $\bullet$ & $\circ$ & $\circ$ & $\circ$ & $\bullet$ & $\circ$ & $\circ$ &$\bullet$  &$\bullet$  & $\circ$ & $\circ$ & $\bullet$ & $\bullet$ & $\bullet$ & $\bullet$ &                            \\ 
\hline
\multirow{12}{*}{1\%} & \multirow{4}{*}{ET} & RFNet & 7.39 & 42.96 & 8.75 & 13.09 & 38.65 & 44.62 & 15.59 & 8.71 & 13.40 & 39.86 & 35.96 & 13.76 & 30.76 & 32.18 & 27.47 & 24.88 \\
 &  & M3AE & 40.05 & 66.67 & 36.13 & 36.15 & 72.06 & 66.88 & 38.15 & 42.18 & 40.78 & 68.47 & 72.15 & 37.80 & 71.51 & 70.51 & 73.53 & 55.53 \\
 &  & Meta & 25.32 & 58.33 & 20.03 & 22.71 & 65.65 & 64.49 & 21.01 & 26.86 & 24.80 & 67.12 & 67.40 & 25.37 & 69.08 & 69.26 & 67.55 & 46.33 \\
 &  & Ours & \textbf{43.74} & \textbf{74.82} & \textbf{38.92} & \textbf{41.01} & \textbf{75.88} & \textbf{75.48} & \textbf{42.67} & \textbf{44.81} & \textbf{45.00} & \textbf{75.52} & \textbf{75.10} & \textbf{45.78} & \textbf{75.67} & \textbf{75.96} & \textbf{75.59} & \textbf{60.40} \\ \cline{2-19} 
 & \multirow{4}{*}{TC} & RFNet & 16.48 & 42.15 & 28.12 & 17.66 & 35.35 & 42.92 & 20.65 & 22.21 & 17.96 & 33.13 & 33.39 & 19.07 & 29.90 & 34.31 & 28.10 & 28.09 \\
 &  & M3AE & 67.60 & 79.93 & 65.53 & 65.35 & 81.56 & 80.43 & 69.87 & 69.63 & 70.70 & \textbf{83.28} & \textbf{82.62} & 70.30 & 82.24 & 82.31 & 81.52 & 75.53 \\
 &  & Meta & 50.00 & 70.87 & 47.04 & 47.83 & 75.19 & 71.96 & 52.64 & 54.57 & 52.33 & 74.84 & 73.98 & 57.49 & 74.99 & 76.41 & 74.79 & 63.66 \\
 &  & Ours & \textbf{71.84} & \textbf{82.81} & \textbf{66.82} & \textbf{70.01} & \textbf{83.42} & \textbf{83.18} & \textbf{71.03} & \textbf{73.70} & \textbf{73.88} & 82.65 & 82.56 & \textbf{74.73} & \textbf{83.30} & \textbf{83.56} & \textbf{83.19} & \textbf{77.78} \\ \cline{2-19} 
 & \multirow{4}{*}{WT} & RFNet & 61.36 & 46.45 & 46.19 & 66.43 & 62.95 & 50.67 & 70.16 & 62.99 & 73.33 & 66.38 & 67.42 & 74.23 & 72.46 & 62.85 & 72.19 & 63.74 \\
 &  & M3AE & 84.61 & 75.99 & \textbf{76.83} & 87.76 & 84.07 & \textbf{77.27} & 88.58 & 84.16 & 89.02 & \textbf{89.07} & \textbf{88.59} & 88.65 & 88.02 & 84.26 & 87.68 & 84.97 \\
 &  & Meta & 77.12 & 66.90 & 62.65 & 82.15 & 77.59 & 70.82 & 82.24 & 74.69 & 83.46 & 83.61 & 83.14 & 84.69 & 84.32 & 77.66 & 83.15 & 78.28 \\
 &  & Ours & \textbf{85.45} & \textbf{76.10} & 75.12 & \textbf{88.31} & \textbf{84.90} & 77.26 & \textbf{88.78} & \textbf{86.08} & \textbf{89.60} & 87.99 & 88.06 & \textbf{89.69} & \textbf{88.84} & \textbf{85.14} & \textbf{88.90} & \textbf{85.35} \\ 
 \hline\hline
\multirow{12}{*}{10\%} & \multirow{4}{*}{ET} & RFNet & 29.15 & 62.35 & 17.64 & 23.39 & 68.85 & 68.69 & 25.32 & 29.64 & 25.80 & 71.64 & 70.72 & 30.23 & 70.45 & 70.91 & 70.39 & 49.01 \\
 &  & M3AE & 43.18 & 73.62 & \textbf{39.55} & 37.70 & 74.34 & 73.04 & 39.42 & 44.91 & 45.38 & \textbf{75.42} & \textbf{73.28} & 45.53 & 74.43 & \textbf{74.83} & 75.23 & 59.32 \\
 &  & Meta & 42.40 & 65.91 & 39.19 & \textbf{38.77} & 66.86 & 64.57 & 38.31 & 41.00 & 42.56 & 66.33 & 67.89 & 41.91 & 66.99 & 67.77 & 69.13 & 54.64 \\
 &  & Ours & \textbf{46.57} & \textbf{74.50} & 38.74 & 38.39 & \textbf{74.53} & \textbf{73.77} & \textbf{42.27} & \textbf{46.99} & \textbf{45.93} & 74.68 & 73.25 & \textbf{48.22} & \textbf{75.20} & 74.06 & \textbf{75.39} & \textbf{60.17} \\ \cline{2-19} 
 & \multirow{4}{*}{TC} & RFNet & 48.88 & 70.10 & 46.06 & 48.58 & 75.91 & 73.22 & 53.50 & 55.09 & 53.59 & 73.51 & 74.26 & 56.34 & 74.48 & 76.31 & 74.63 & 63.63 \\
 &  & M3AE & 69.19 & 81.00 & 66.54 & 67.31 & 82.31 & \textbf{81.50} & 70.02 & 71.04 & 71.31 & 82.23 & 81.56 & 72.69 & 82.37 & \textbf{81.97} & 82.13 & 76.21 \\
 &  & Meta & 63.39 & 75.70 & 59.91 & 60.68 & 75.11 & 74.76 & 64.91 & 64.58 & 65.83 & 78.35 & 78.37 & 67.25 & 76.96 & 76.10 & 77.48 & 70.63 \\
 &  & Ours & \textbf{72.77} & \textbf{81.20} & \textbf{68.52} & \textbf{67.53} & \textbf{82.95} & 80.90 & \textbf{71.50} & \textbf{72.61} & \textbf{73.00} & \textbf{83.28} & \textbf{83.58} & \textbf{74.03} & \textbf{83.62} & 81.49 & \textbf{83.86} & \textbf{77.39} \\ \cline{2-19} 
 & \multirow{4}{*}{WT} & RFNet & 79.98 & 66.82 & 68.11 & 80.01 & 82.75 & 72.32 & 83.55 & 82.24 & 85.23 & 85.19 & 85.33 & 86.15 & 86.80 & 82.95 & 86.82 & 80.95 \\
 &  & M3AE & 84.58 & 75.68 & \textbf{77.08} & \textbf{88.28} & 84.12 & 77.06 & 87.84 & \textbf{84.36} & 88.19 & \textbf{89.07} & 88.10 & 88.05 & 88.28 & 84.20 & 88.25 & 84.88 \\
 &  & Meta & 77.77 & 71.84 & 71.53 & 82.50 & 78.45 & 74.33 & 83.94 & 79.11 & 85.20 & 83.62 & 82.78 & 85.59 & 84.50 & 76.84 & 83.67 & 80.11 \\
 &  & Ours & \textbf{85.88} & \textbf{76.54} & 75.78 & 88.11 & \textbf{85.64} & \textbf{78.01} & \textbf{88.13} & 84.34 & \textbf{89.25} & 88.43 & \textbf{88.70} & \textbf{89.12} & \textbf{89.02} & \textbf{85.04} & \textbf{89.10} & \textbf{85.41} \\ 
 \hline\hline
\multirow{12}{*}{50\%} & \multirow{4}{*}{ET} & RFNet & 41.11 & 66.00 & 25.47 & 30.95 & 72.60 & 69.50 & 37.17 & 41.58 & 41.11 & 72.23 & 72.90 & 43.26 & 72.64 & 72.22 & 73.39 & 55.48 \\
 &  & M3AE & 42.73 & 73.68 & 36.77 & 36.44 & 75.79 & 74.25 & 39.94 & 45.51 & 43.82 & \textbf{74.06} & 74.26 & 44.66 & 75.61 & \textbf{75.95} & 74.22 & 59.18 \\
 &  & Meta & 43.28 & 72.14 & 37.43 & 39.13 & 72.72 & 72.26 & 42.40 & 46.64 & 45.65 & 72.22 & 73.76 & 44.16 & 74.31 & 72.73 & 73.49 & 58.82 \\
 &  & Ours & \textbf{48.73} & \textbf{75.52} & \textbf{37.56} & \textbf{40.22} & \textbf{76.05} & \textbf{75.84} & \textbf{44.27} & \textbf{48.94} & \textbf{46.29} & 73.99 & \textbf{74.48} & \textbf{48.50} & \textbf{75.66} & 75.86 & \textbf{76.09} & \textbf{61.20} \\ \cline{2-19} 
 & \multirow{4}{*}{TC} & RFNet & 60.73 & 76.34 & 59.05 & 54.18 & 80.63 & 77.99 & 68.80 & 67.06 & 64.31 & 80.85 & 80.56 & 69.23 & 80.95 & 80.49 & 80.88 & 72.14 \\
 &  & M3AE & 70.51 & \textbf{83.64} & 65.90 & 66.59 & 84.10 & \textbf{83.79} & 69.40 & 70.82 & 72.26 & \textbf{85.26} & \textbf{84.47} & 72.47 & 84.42 & \textbf{84.49} & 84.05 & 77.48 \\
 &  & Meta & 71.30 & 81.54 & \textbf{68.17} & \textbf{69.61} & \textbf{84.68} & 82.39 & 71.37 & 71.69 & 72.13 & 83.71 & 82.79 & 72.56 & 84.07 & 83.59 & 83.38 & 77.53 \\
 &  & Ours & \textbf{72.79} & 82.63 & 67.65 & 68.84 & 84.42 & 82.85 & \textbf{71.78} & \textbf{74.29} & \textbf{73.84} & 84.32 & 83.86 & \textbf{74.10} & \textbf{84.62} & 83.92 & \textbf{84.42} & \textbf{78.29} \\ \cline{2-19} 
 & \multirow{4}{*}{WT} & RFNet & 84.22 & 71.52 & 73.08 & 84.00 & 85.34 & 76.51 & 87.44 & 85.05 & 88.06 & 88.09 & 88.49 & 88.86 & 89.14 & 85.57 & 89.19 & 84.30 \\
 &  & M3AE & 84.47 & 76.29 & 75.50 & 88.52 & 84.47 & 77.64 & 88.58 & 84.26 & \textbf{89.63} & 89.14 & 89.13 & 89.15 & 89.55 & 84.69 & 88.24 & 85.28 \\
 &  & Meta & \textbf{85.64} & 76.18 & 76.58 & 88.48 & 85.72 & 77.25 & 89.05 & 84.68 & 89.20 & 88.68 & 88.81 & \textbf{89.22} & 88.58 & 84.22 & 88.26 & 85.37 \\
 &  & Ours & 85.50 & \textbf{77.80} & \textbf{76.77} & \textbf{88.64} & \textbf{86.16} & \textbf{79.11} & \textbf{89.24} & \textbf{85.35} & 89.24 & \textbf{89.79} & \textbf{89.67} & 89.21 & \textbf{90.05} & \textbf{86.07} & \textbf{89.96} & \textbf{86.17} \\
 \hline\hline
\end{tabular}
}
\label{tab:main_brats18}
\end{table}

\subsection{Implementation Details}
\label{ssec:implementation}

The experiments are implemented with Pytorch using two NVIDIA RTX A5000 GPUs. We employ a 3D U-Net with group normalization~\cite{wu2018groupnorm} used in previous works~\cite{wang2021acn, liu2023m3ae} as our backbone. To construct the training set $\mathcal{D} = \mathcal{D}_f + \mathcal{D}_m$, we randomly select subjects from the train set according to the full modality (FM) ratio to compose $\mathcal{D}_f$. The remaining samples, after randomly discarding 1 to 3 modalities, constitute $\mathcal{D}_m$. For data augmentation, we first randomly crop the input images to $128 \times 128 \times 128$, and then perform min-max scaling to standardize all the volumes. Other commonly-used augmentation like random intensity shifts, random intensity scaling, and random flipping with a probability of 0.5 are also conducted. The Adam optimizer with an initial learning rate of 3e-4 is utilized for both pre-training and segmentation model training. The weight decay is set to 5e-4. The learning rate is adjusted during training by a cosine decay scheduler~\cite{loshchilov2017sgdr}. For pre-training, we set the batch size to 2 and train for 600 epochs. For segmentation model training stage, early stopping is employed to prevent overfitting, training for a maximum of 1000 epochs with a batch size of 1. At the inference stage, only the segmentation model (student) is retained for evaluating performance under 15 different modality-input scenarios. 

\begin{table}[t]
\centering
\caption{Comparison with state-of-the-art on segmentation of tumor regions (ET, TC, WT) under different FM ratios on BraTS2020. The average DSC across 15 input modality scenarios are reported.}
\scalebox{0.75}{
\begin{tabular}{c|l|cccc|c|l|cccc}
\hline\hline
\multirow{2}{*}{FM Ratio} & \multicolumn{1}{c|}{\multirow{2}{*}{Methods}} & \multicolumn{4}{c|}{Average DSC (\%)} & \multirow{2}{*}{FM Ratio} & \multicolumn{1}{c|}{\multirow{2}{*}{Methods}} & \multicolumn{4}{c}{Average DSC (\%)} \\ \cline{3-6} \cline{9-12} 
 & \multicolumn{1}{c|}{} & ET & TC & WT & Avg. &  & \multicolumn{1}{c|}{} & ET & TC & WT & Avg. \\ \hline
\multirow{4}{*}{1\% FM} & RFNet & 28.87 & 38.14 & 66.55 & 44.52 & \multirow{4}{*}{50\% FM} & RFNet & 61.21 & 76.64 & 85.67 & 74.51 \\
 & M3AE & 61.15 & 78.06 & 86.20 & 75.14 &  & M3AE & 62.37 & 79.01 & 86.61 & 76.00 \\
 & Meta & 47.29 & 64.17 & 78.79 & 63.42 &  & Meta & 62.44 & 78.30 & 86.09 & 75.61 \\
 & Ours & \textbf{63.07} & \textbf{78.54} & \textbf{86.79} & \textbf{76.13} &  & Ours & \textbf{64.59} & \textbf{79.23} & \textbf{87.14} & \textbf{76.99} \\ \hline
\multirow{4}{*}{10\% FM} & RFNet & 55.31 & 61.71 & 82.90 & 66.64 & \multirow{4}{*}{100\% FM} & RFNet & 62.08 & 77.97 & 86.33 & 75.46 \\
 & M3AE & 62.45 & 78.88 & 86.19 & 75.84 &  & M3AE & 63.52 & 79.12 & 86.72 & 76.45 \\
 & Meta & 57.11 & 70.07 & 83.02 & 70.07 &  & Meta & - & - & - & - \\
 & Ours & \textbf{63.96} & \textbf{78.99} & \textbf{86.53} & \textbf{76.49} &  & Ours & \textbf{64.25} & \textbf{79.80} & \textbf{87.41} & \textbf{77.15} \\ \hline\hline
\end{tabular}
}
\label{tab:main_brats20}
\end{table}

\subsection{Comparison with State-of-the-arts}
\label{ssec:comp_with_sota}

We compare our method with SOTA approaches including RFNet~\cite{ding2021rfnet}, M$^3$AE~\cite{liu2023m3ae}, and Meta-learning~\cite{konwer2023enhancing} on BraTS2018 and BraTS2020. RFNet is representative of traditional methods, which focus on missing modality in the inference stage while relying on full modality in training stage. 
M$^3$AE, although originally designed to address missing modalities during the inference phase as well, has a flexible structure that allows modality incompleteness during training. 
In implementation, we make a straightforward extension to allow it to adapt to different FM ratios in training. 
Meta-learning is the most related work which is designed for dealing with partial modality training data with 50\% FM ratio.
However, given that the code of Meta-learning is unavailable, we instead implement their framework by ourselves following the technical details in the paper. To ensure a fair comparison, all the methods have the same 3D U-Net backbone.
We train these missing modality methods using training sets with varying FM ratios (1\%, 10\%, and 50\%), then perform inference under 15 different missing modality scenarios, reporting the average DSC and HD95 separately for ET, TC, WT, and the mean of three tumor areas. 
Note that as these approaches are not originally designed for varying FM ratios of training data and with different dataset splits, we try our best to keep their design principle and adapt them to our task. All these approaches are implemented with the same dataset split and network backbone as our proposed method, for fair comparison.
As shown in Tab.~\ref{tab:main_brats18} and Tab.~\ref{tab:main_brats20}, our approach outperforms SOTA methods under any FM ratio, demonstrating our robustness to missing modalities during the training phase. As traditional methods, such as RFNet, can only accept full modality data during training, the number of available subjects decreases along with the FM ratio, leading to a significant performance decline. Both meta-learning and the extended M$^3$AE can leverage subjects in $\mathcal{D}_f$ and $\mathcal{D}_m$. However, due to the requirement for sufficient full modality data for meta-test, Meta-learning struggles to effectively address the issue when the missing modality problem at the training stage becomes severe (10\% and 1\% FM). Our method demonstrates performance surpassing M$^3$AE across various FM ratios, proving that our framework can effectively utilize the recovered missing modality information, which goes beyond the acceptance of incomplete training modalities.

\begin{figure}[h]
	\centering
		\begin{minipage}{0.41\linewidth}
			\centering
            \tabcaption{Comparison with other missing modality methods trained with 100\% FM data on BraTS2018. Average DSC across 15 modality combinations are reported.}
            \scalebox{0.8}{
			\begin{tabular}{l|cccc}
            \hline
            \multicolumn{1}{c|}{\multirow{2}{*}{Methods}} & \multicolumn{4}{c}{Average DSC (\%)} \\ \cline{2-5} 
            \multicolumn{1}{c|}{} & ET & TC & WT & Avg. \\ \hline
            HeMIS & 40.79 & 60.67 & 78.15 & 59.87 \\
            U-HVED & 41.12 & 61.19 & 78.71 & 60.34 \\
            ACN & 54.22 & 73.42 & 83.25 & 70.30 \\
            SMU-Net & 55.05 & 72.31 & 83.34 & 70.23 \\
            RFNet & 56.43 & 75.17 & 85.31 & 72.30 \\
            M3AE & 60.00 & 76.98 & 85.33 & 74.10 \\ \hline
            Ours 1\% & 60.41 & 77.77 & 85.32 & 74.50 \\
            Ours 100\% & \textbf{61.31} & \textbf{78.91} & \textbf{85.69} & \textbf{75.30} \\ \hline
            \end{tabular}}
			\label{tab:comp_w_other_sota}
		\end{minipage}
	\hspace{15pt}
		\begin{minipage}[h]{0.41\linewidth}
			\centering
			\includegraphics[width=1.0\linewidth]{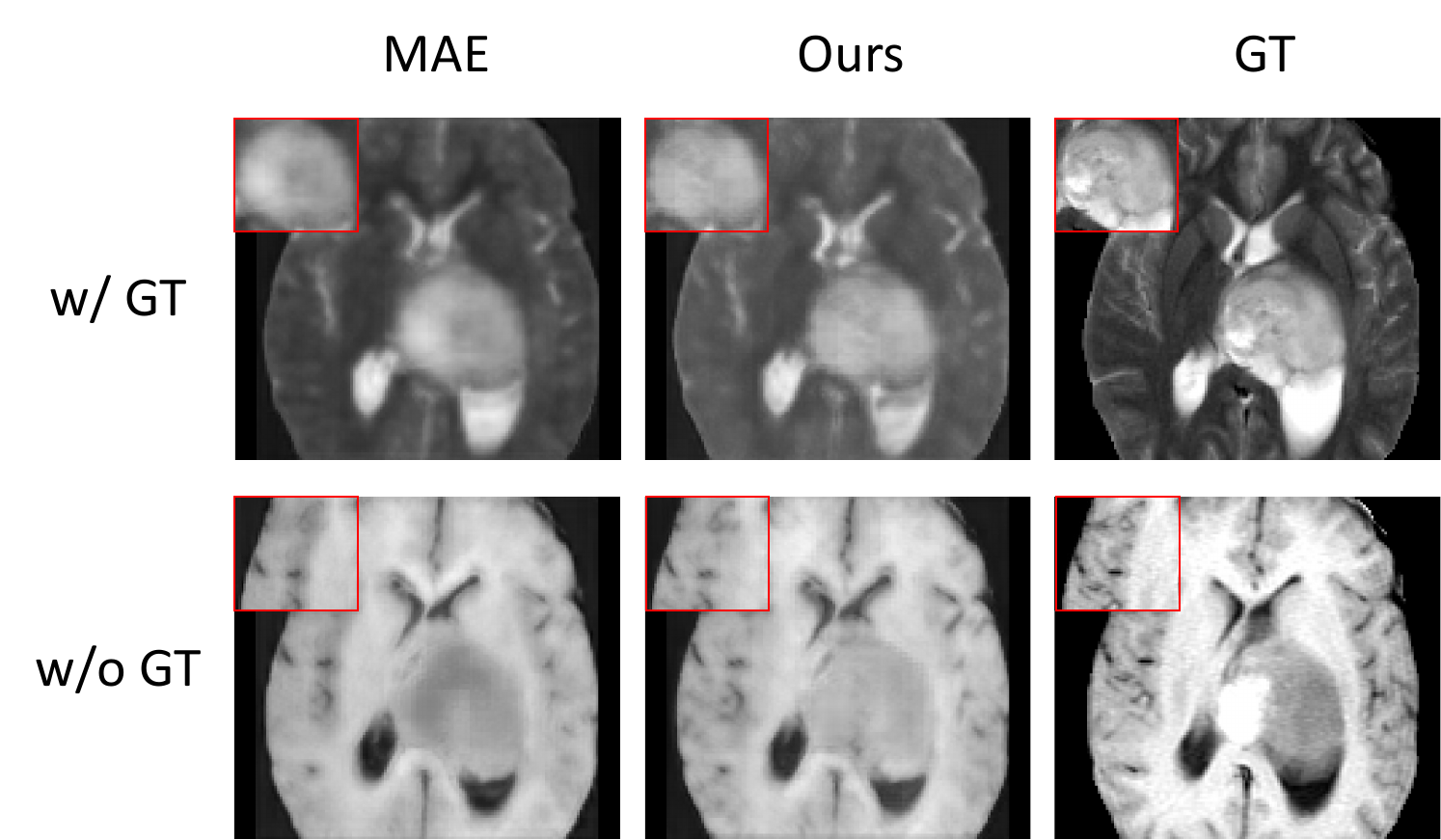}
	        \figcaption{Visualization of reconstructed training modalities. 
            Row 1 indicates that ground truth (GT) is available during pre-training, while row 2 means GT is absent. Column 1-3 represents vanilla multimodal MAE, ours, and GT.
            }
	        \label{fig:vis_inv}
		\end{minipage}
\end{figure}

As shown in Tab.~\ref{tab:comp_w_other_sota}, we also compare our framework with other SOTA methods for the traditional missing modality setting including HeMIS~\cite{havaei2016hemis}, U-HVED~\cite{dorent2019hved}, ACN~\cite{wang2021acn}, and SMU-Net~\cite{azad2022smunet}. All these methods are trained with 100\% full modality data and the results are quoted from~\cite{liu2023m3ae}, except HeMIS, RFNet and M$^3$AE itself.
It can be seen that our method trained with 1\% FM data achieves comparable results to others, and surpasses others when all the training data are modality-complete, \ie, under 100\% FM ratio. The comparison results reveal our superiority in dealing with missing modalities during both training and the inference phase.

\subsection{Ablation Studies}
\label{ssec:ablation}

\textbf{Effectiveness of key components.} Several ablations are performed to evaluate the effectiveness of each proposed module in our framework. We adopt a multimodal MAE as our baseline. Specifically, for modality-incomplete samples during pre-training, we calculate the reconstruction loss only on available modalities. Modality dropout is used to fine-tune the missing modality segmentation model. From Tab.~\ref{tab:ablation}, it can be seen that our modality-incomplete pre-training strategy outperforms the baseline by 1.43\% in the average of three tumor regions. This proves that our distribution approximation can provide auxiliary supervision to compensate for the incomplete self-supervision caused by missing modalities during training, resulting in high-quality pre-trained weights. Based on this, we further incorporate the CLIP-driven hyper-network to our segmentation model during training, aiming to enhance its representation capability for heterogeneous modality inputs during both training and inference stages, resulting in a 2.31\% improvement compared to the baseline. We also utilize data-model co-distillation to effectively transfer the recovered information of absent modalities, and finally achieved a 2.49\% advantage relative to baseline.

\begin{table}[ht]
\centering
\caption{Ablation study of key components.}
\scalebox{0.9}{
\begin{tabular}{cl|cccc}
\hline
\multicolumn{2}{c|}{\multirow{2}{*}{Methods}}                                  & \multicolumn{4}{c}{Average DSC (\%)} \\ 
\cline{3-6} 
\multicolumn{2}{c|}{}                                                          & ET      & TC      & WT      & Avg.   \\ 
\hline
\multicolumn{1}{c|}{\multirow{4}{*}{1\% FM}}   & MAE baseline                          & 55.53   & 75.53   & 84.97   & 72.01  \\
\multicolumn{1}{c|}{}                          & $+$distribution approximation   & 58.34   & 76.99   & 84.99   & 73.44  \\
\multicolumn{1}{c|}{}                          & $+$CLIP-driven hypernetwork   & 59.76   & 77.68   & \textbf{85.51}   & 74.32  \\
\multicolumn{1}{c|}{}                          & $+$data-model co-distillation & \textbf{60.41}   &  \textbf{77.77}   &  85.32   &  \textbf{74.50}  \\ 
\hline
\end{tabular}
}
\label{tab:ablation}
\end{table}

\begin{table}[b]
\footnotesize
\centering
\caption{Ablation study of data-model co-distillation. We use binary modality codes as indicators of hyper-network for experiments in this table to better demonstrate the effectiveness of components in data-model co-distillation.}
\scalebox{0.95}{
\begin{tabular}{ll|c|c|c|c|cccc}
\hline
\multicolumn{2}{l|}{\multirow{2}{*}{Methods}} & \multirow{2}{*}{MAE} & \multirow{2}{*}{KD} & \multirow{2}{*}{\begin{tabular}[c]{@{}c@{}}Data \\ Refine\end{tabular}} & \multirow{2}{*}{Hyp-T} & \multicolumn{4}{c}{Average DSC (\%)} \\ \cline{7-10} 
\multicolumn{2}{l|}{} &  &  &  &  & ET & TC & WT & Avg. \\ \hline
\multicolumn{2}{l|}{(a) distribution approximation} & \multicolumn{1}{l|}{} & \multicolumn{1}{l|}{} & \multicolumn{1}{l|}{} &  & 58.34 & 76.99 & 84.99 & 73.44 \\
\multicolumn{2}{l|}{(b) model dist. w/o MAE} &  & $\checkmark$ &  & $\checkmark$ & 58.35 & 76.79 & 85.09 & 73.41 \\
\multicolumn{2}{l|}{(c) model dist. w/ MAE} & $\checkmark$ & $\checkmark$ &  & $\checkmark$ & 59.21 & 77.21 & 85.17 & 73.87 \\
\multicolumn{2}{l|}{(d) co-dist. w/o hyp-T} & $\checkmark$ & $\checkmark$ & $\checkmark$ &  & 58.86 & 77.41 & 84.98 & 73.75 \\
\multicolumn{2}{l|}{(e) data-model co-dist.} & $\checkmark$ & $\checkmark$ & $\checkmark$ & $\checkmark$ & 59.61 & 78.10 & 85.05 & 74.25 \\ \hline
\end{tabular}
}
\label{tab:dmcd}
\end{table}

\begin{figure}[ht]
\begin{minipage}[b]{.28\linewidth}
  \centering
  \includegraphics[width=1.0\linewidth]{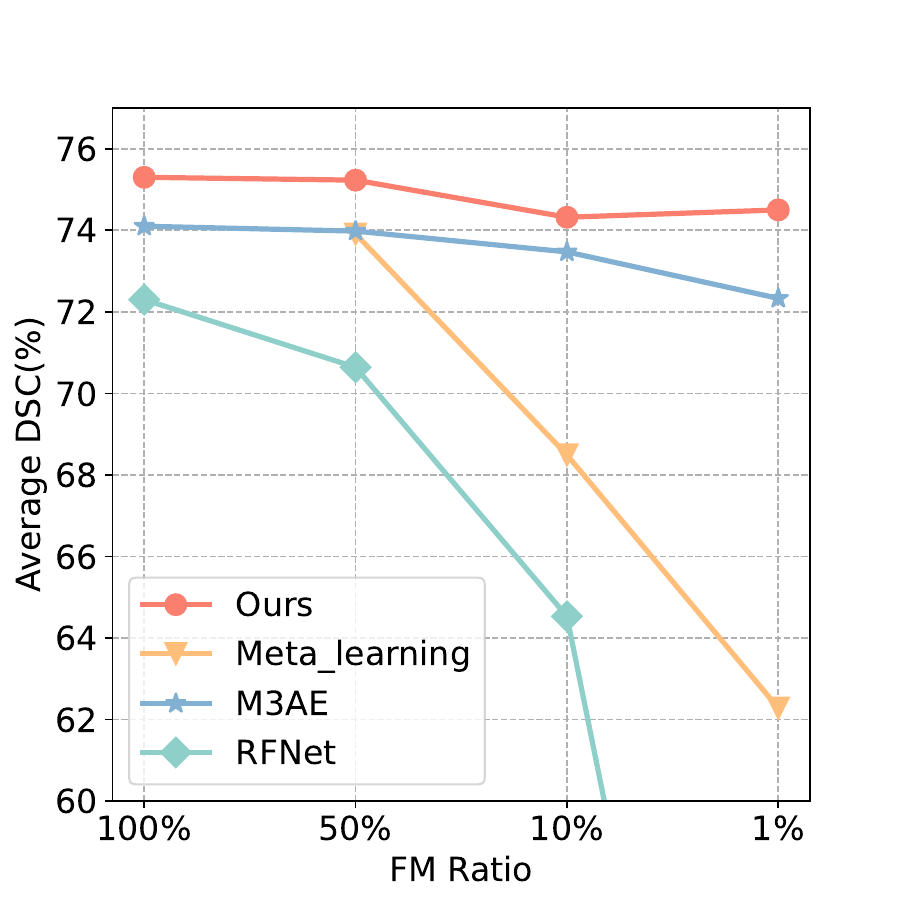}
  \caption{Results for varying full modality ratio in training on BraTS2018.}
  \label{fig:fm_ratio}
\end{minipage}
\begin{minipage}[b]{.69\linewidth}
  \centering
  \includegraphics[width=1.0\linewidth]{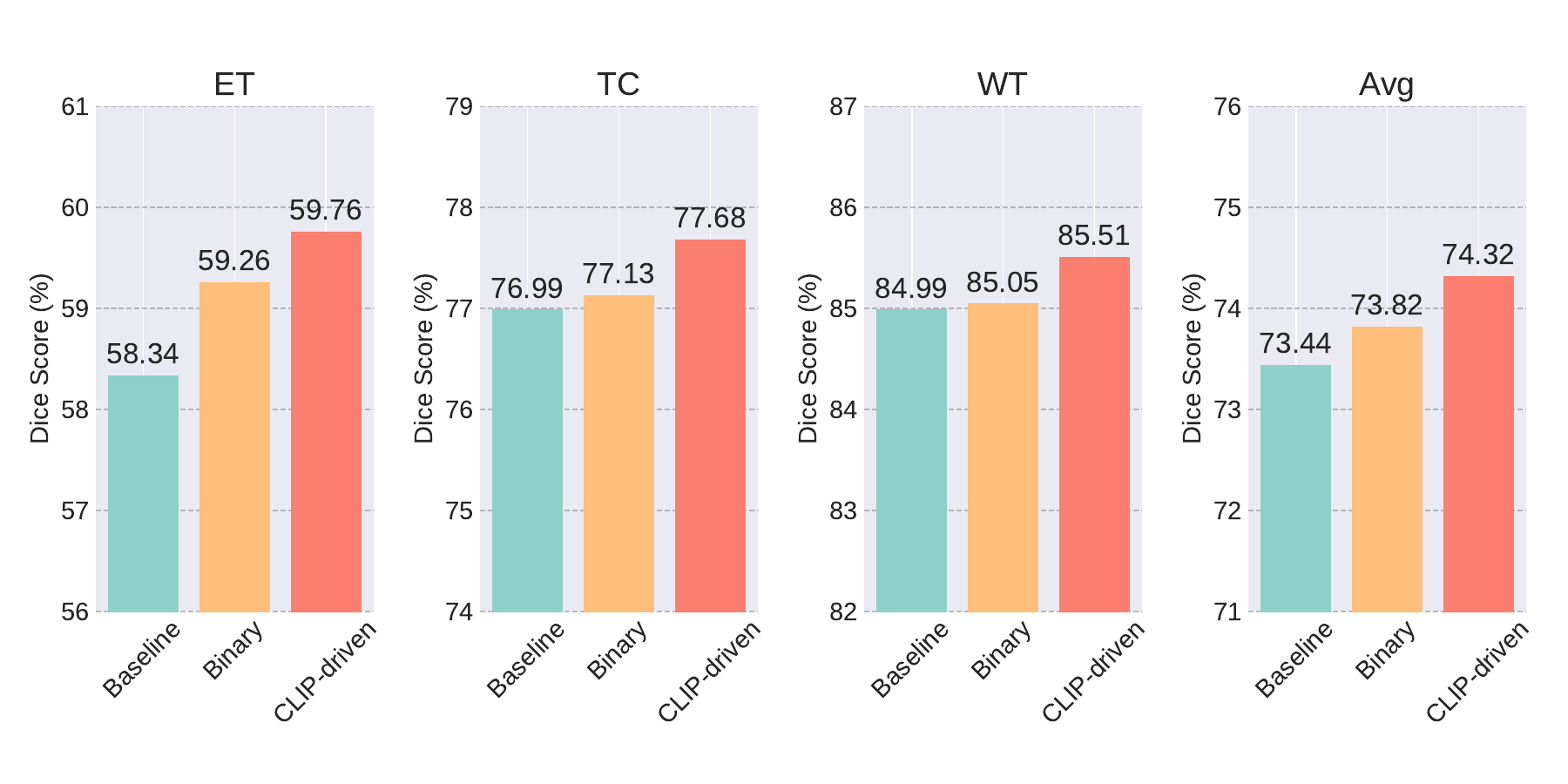}
  \caption{Ablation study of CLIP-dirven hyper-network.}
  \label{fig:hyp_ablation}
\end{minipage}
\end{figure}

\begin{figure}[ht]
  \centering
  \includegraphics[width=0.9\linewidth]{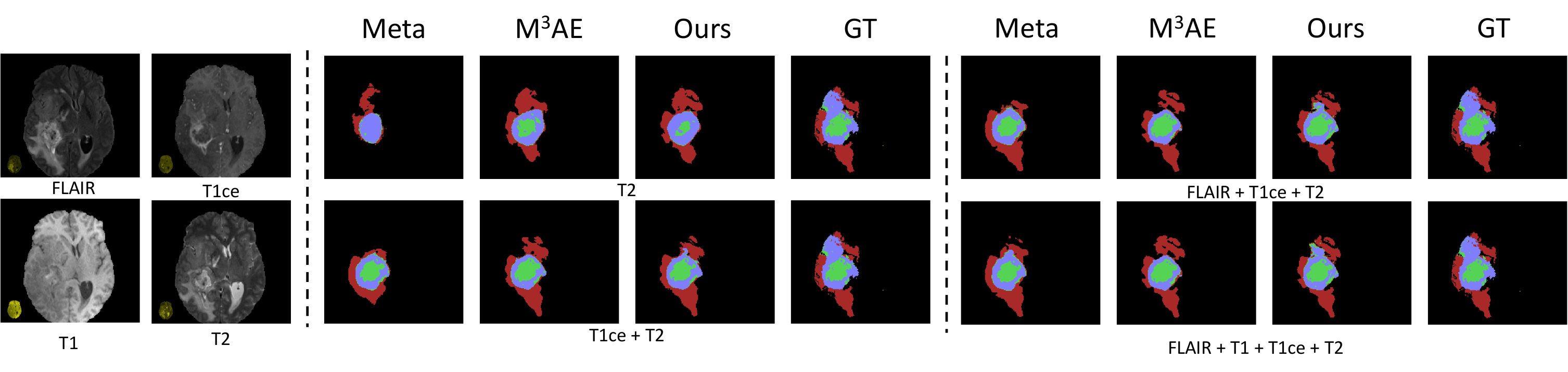}
  \caption{Visualization results from 4 methods for different combinations of modalities. All the methods are trained under the 1\% FM setting.}
  \label{fig:seg_vis}
\end{figure}

\noindent\textbf{Detail analysis of data-model co-distillation.} We utilize modality drop to train the segmentation model initialized by weights from our distribution approximation pre-training strategy, thereby formulating the baseline. The segmentation model is not equipped with CLIP-driven hyper-network to better highlight the differences between experimental results. Experiments are shown in Tab.~\ref{tab:dmcd}. 
(b) and (c) use a fixed teacher model for knowledge distillation. Wherein, (b) indicates knowledge transfer from all available modalities to modality-dropped representations in our segmentation model. (c), on the other hand, utilizes the pre-trained MAE to additionally synthesize information for missing modalities before distilling it to the segmentation model. 
(d) and (e) indicate the joint optimization of MAE-synthesized missing modality data during the model distillation process, with the distinction that (d) did not employ a hyper-teacher.
Comparing (a), (b), and (c), we find that it is effective to adopt KD to transfer the recovered modality knowledge, while KD itself without synthesizing missing modality information is useless. From (c) vs. (e), we empirically demonstrate that the recovered modalities should be further refined along with KD to be more informative. Finally, from (d) vs. (e), we prove that hyper-teacher is crucial to our data-model co-distillation, as it can provide unique full modality knowledge by setting the teacher model to full modality mode.

\noindent\textbf{Detail analysis of CLIP-driven hyper-network.} Fig.~\ref{fig:hyp_ablation} shows the ablation results of our CLIP-driven hyper-network. The baseline remains the same as in Tab.~\ref{tab:dmcd}. Experiments are conducted under 1\% FM ratio, and average DSCs across 15 modality combinations are reported. We replace the CLIP text encoder with an MLP to generate modality embedding using binary modality code, \eg,~[1, 0, 0, 1] for a sample with FLAIR, and T2. It can be seen that an improvement of 0.38\% on the mean of three regions is achieved by using hyper-network with binary indicator. Nevertheless, binary codes assume that each modality is independent to other modalities and ignore the inter-modal correlations. 
After using a frozen CLIP encoder to extract textual embedding from MRI modality prompts, the performance on all three classes improves, and the average increases from 73.82\% to 74.32\%.

\noindent\textbf{Visual comparison of reconstructed modalities.} Fig.~\ref{fig:vis_inv} visually compares the reconstructed results before and after using our distribution approximation. The ground truth in the first row is available for self-supervision. After providing extra supervision by distribution approximation, the tumor area of reconstructed images is slightly clearer. This demonstrates that auxiliary supervision can encourage the pre-training process to focus more on tumor-aware reconstruction. From the second row where the ground truth is absent, we see that more brain anatomical details are recovered, proving effectiveness of auxiliary supervision.

\section{Conclusion}
In this paper, we present a universal model to tackle missing modality in both training and inference phases. 
We innovate a distribution approximation mechanism to pre-train our model for modality reconstruction. Based on the reconstructed full modality, we propose a data-model co-distillation scheme to leverage its inherent knowledge, in which we design a CLIP-driven hyper-network to alleviate distribution heterogeneity.
By doing robust modality reconstruction and model personalization, our universal model demonstrates superior performance on two brain tumor segmentation benchmarks, surpassing other methods under all missing phases with various missing ratios.

\bibliographystyle{splncs04}
\bibliography{ref}
\end{document}